\documentclass{article}

\usepackage[nonatbib,preprint]{neurips_2022}

\usepackage[utf8]{inputenc} % allow utf-8 input
\usepackage[T1]{fontenc}    % use 8-bit T1 fonts
\usepackage{hyperref}       % hyperlinks
\usepackage{url}            % simple URL typesetting
\usepackage{booktabs}       % professional-quality tables
\usepackage{amsfonts}       % blackboard math symbols
\usepackage{nicefrac}       % compact symbols for 1/2, etc.
\usepackage{microtype}      % microtypography
\usepackage{xcolor}         % colors
\usepackage{verbatim}

\usepackage{graphicx}
\usepackage{booktabs}
\usepackage{shadowtext}
\usepackage{amssymb}
\usepackage{comment}
\usepackage{amsmath,amssymb} % define this before the line numbering.
\usepackage{sidecap}
\usepackage{color}
\usepackage{amsmath}
\usepackage{caption} 
\usepackage{algorithm,algorithmic}
\usepackage{booktabs}
\usepackage{multirow}
\usepackage{wrapfig}
\usepackage{subfig}
\usepackage{xcolor}
\usepackage{authblk}

\title{Landscape Learning for Optimization-Based Inference}

\title{Landscape Learning for Neural Network Inversion}

\usepackage{soul}
\author[1]{\href{rliu@cs.columbia.edu}{\textbf{Ruoshi Liu}}}
\author[1]{\textbf{Chengzhi Mao}}
\author[1]{\textbf{Purva Tendulkar}}
\author[2]{\textbf{Hao Wang}}
\author[1]{\textbf{Carl Vondrick}}
\affil[1]{Columbia University}
\affil[2]{Rutgers University}

\begin{document}

\maketitle

\def\Blue{\color{blue}}
\def\Purple{\color{purple}}

\def\A{{\bf A}}
\def\a{{\bf a}}
\def\B{{\bf B}}
\def\b{{\bf b}}
\def\C{{\bf C}}
\def\c{{\bf c}}
\def\D{{\bf D}}
\def\d{{\bf d}}
\def\E{{\bf E}}
\def\e{{\bf e}}
\def\f{{\bf f}}
\def\F{{\bf F}}
\def\K{{\bf K}}
\def\k{{\bf k}}
\def\L{{\bf L}}
\def\H{{\bf H}}
\def\h{{\bf h}}
\def\G{{\bf G}}
\def\g{{\bf g}}
\def\I{{\bf I}}
\def\R{{\bf R}}
\def\X{{\bf X}}
\def\Y{{\bf Y}}
\def\OO{{\bf O}}
\def\oo{{\bf o}}
\def\P{{\bf P}}
\def\Q{{\bf Q}}
\def\r{{\bf r}}
\def\s{{\bf s}}
\def\S{{\bf S}}
\def\t{{\bf t}}
\def\T{{\bf T}}
\def\x{{\bf x}}
\def\y{{\bf y}}
\def\z{{\bf z}}
\def\Z{{\bf Z}}
\def\M{{\bf M}}
\def\m{{\bf m}}
\def\n{{\bf n}}
\def\U{{\bf U}}
\def\u{{\bf u}}
\def\V{{\bf V}}
\def\v{{\bf v}}
\def\W{{\bf W}}
\def\w{{\bf w}}
\def\0{{\bf 0}}
\def\1{{\bf 1}}
\def\N{{\bf N}}

\def\AM{{\mathcal A}}
\def\EM{{\mathcal E}}
\def\FM{{\mathcal F}}
\def\TM{{\mathcal T}}
\def\UM{{\mathcal U}}
\def\XM{{\mathcal X}}
\def\YM{{\mathcal Y}}
\def\NM{{\mathcal N}}
\def\OM{{\mathcal O}}
\def\IM{{\mathcal I}}
\def\GM{{\mathcal G}}
\def\PM{{\mathcal P}}
\def\LM{{\mathcal L}}
\def\MM{{\mathcal M}}
\def\DM{{\mathcal D}}
\def\SM{{\mathcal S}}
\def\RB{{\mathbb R}}
\def\EB{{\mathbb E}}

\def\tx{\tilde{\bf x}}
\def\ty{\tilde{\bf y}}
\def\tz{\tilde{\bf z}}
\def\hd{\hat{d}}
\def\HD{\hat{\bf D}}
\def\hx{\hat{\bf x}}
\def\hR{\hat{R}}

\def\Ome{\mbox{\boldmath$\omega$\unboldmath}}
\def\bet{\mbox{\boldmath$\beta$\unboldmath}}
\def\et{\mbox{\boldmath$\eta$\unboldmath}}
\def\ep{\mbox{\boldmath$\epsilon$\unboldmath}}
\def\ph{\mbox{\boldmath$\phi$\unboldmath}}
\def\Pii{\mbox{\boldmath$\Pi$\unboldmath}}
\def\pii{\mbox{\boldmath$\pi$\unboldmath}}
\def\Ph{\mbox{\boldmath$\Phi$\unboldmath}}
\def\Ps{\mbox{\boldmath$\Psi$\unboldmath}}
\def\pss{\mbox{\boldmath$\psi$\unboldmath}}
\def\tha{\mbox{\boldmath$\theta$\unboldmath}}
\def\Tha{\mbox{\boldmath$\Theta$\unboldmath}}
\def\muu{\mbox{\boldmath$\mu$\unboldmath}}
\def\Si{\mbox{\boldmath$\Sigma$\unboldmath}}
\def\Gam{\mbox{\boldmath$\Gamma$\unboldmath}}
\def\gamm{\mbox{\boldmath$\gamma$\unboldmath}}
\def\Lam{\mbox{\boldmath$\Lambda$\unboldmath}}
\def\De{\mbox{\boldmath$\Delta$\unboldmath}}
\def\vps{\mbox{\boldmath$\varepsilon$\unboldmath}}
\def\Up{\mbox{\boldmath$\Upsilon$\unboldmath}}
\def\Lap{\mbox{\boldmath$\LM$\unboldmath}}
\newcommand{\ti}[1]{\tilde{#1}}

\def\tr{\mathrm{tr}}
\def\etr{\mathrm{etr}}
\def\etal{{\em et al.\/}\,}
\newcommand{\indep}{{\;\bot\!\!\!\!\!\!\bot\;}}
\def\argmax{\mathop{\rm argmax}}
\def\argmin{\mathop{\rm argmin}}
\def\vec{\text{vec}}
\def\cov{\text{cov}}
\def\dg{\text{diag}}

\newcommand{\tabref}[1]{Table~\ref{#1}}
\newcommand{\secref}[1]{Sec.~\ref{#1}}
\newcommand{\figref}[1]{Fig.~\ref{#1}}
\newcommand{\lemref}[1]{Lemma~\ref{#1}}
\newcommand{\thmref}[1]{Theorem~\ref{#1}}
\newcommand{\clmref}[1]{Claim~\ref{#1}}
\newcommand{\crlref}[1]{Corollary~\ref{#1}}
\newcommand{\eqnref}[1]{Eqn.~\ref{#1}}

\newtheorem{remark}{Remark}
\newtheorem{theorem}{Theorem}
\newtheorem{lemma}{Lemma}
\newtheorem{definition}{Definition}

\newtheorem{proposition}{Proposition}

\newcommand{\cz}[1]{{\color{blue}[Chengzhi says: #1]}}
\newcommand{\rl}[1]{{\color{red}[Ruoshi says: #1]}}

\vspace{-1em}
\begin{abstract}
Many machine learning methods operate by inverting a neural network at inference time, which has become a popular technique for solving inverse problems in computer vision, robotics, and graphics.
%which provides a number of advantages to enforce constraints, generate multiple predictions, and remain robust to out-of-distribution data.
However, these methods often involve gradient descent through a highly non-convex loss landscape, causing the optimization process to be unstable and slow. We introduce a method that learns a loss landscape where gradient descent is efficient, bringing massive improvement and acceleration to the inversion process. We demonstrate this advantage on a number of methods for both generative and discriminative tasks, including GAN inversion, adversarial defense, and 3D human pose reconstruction.

\end{abstract}

% We create a new parameter space and project this space to the original parameter space with a projection neural network. By optimizing the parameters in the new space and the projection network with Expectation-Maximization, the projection network learns a smoother loss landscape to perform faster gradient descent. Experimental studies have shown that our framework is widely applicable to various optimization-based inference algorithms and brings massive acceleration in the optimization process. Our framework also generalizes better to OOD data.

\everypar{\looseness=-1}

\section{Introduction}

\begin{wrapfigure}{r}{3in}
    % \centering
    \vspace{-3mm}
    \includegraphics[width=\linewidth]{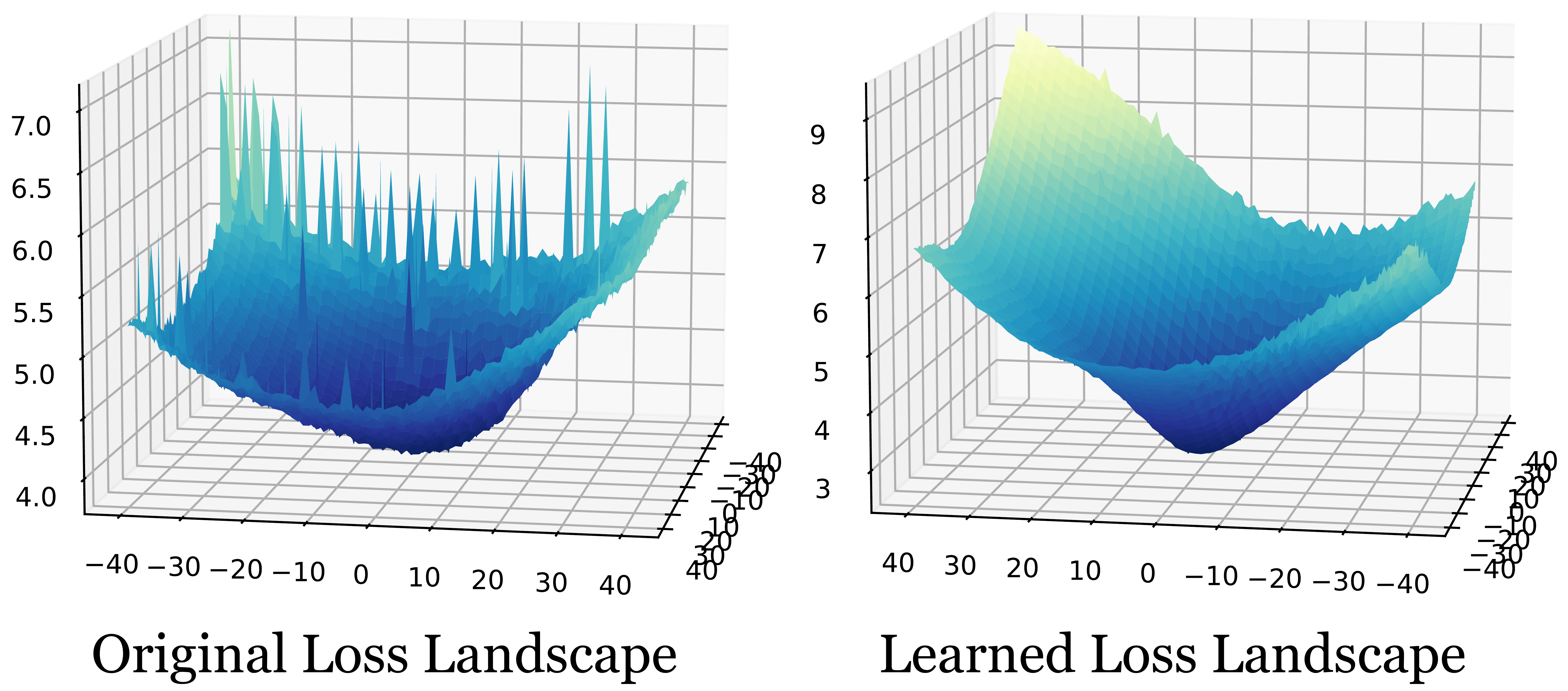}
    \caption{\textbf{Loss Landscapes Comparison.} The loss landscape of optimization-based inference (OBI) is often highly non-convex. We propose to learn a smoother loss landscape through a mapping network to accelerate the optimization procedure. Plotted from real data.}
    \label{fig:teaser1}
\end{wrapfigure}

Many inference problems in machine learning are formulated as inverting a forward model $F(x)$ by optimizing an objective over the input space $x$.
This approach, which we term optimization-based inference (OBI),  has traditionally been used to solve a range of inverse problems in vision, graphics, robotics, recommendation systems, and security \cite{hernandez2008multiview, hernandez2008multiview, lee1993shape, domke2012generic, brakel2013training, stoyanov2011empirical, cremer2019optimization}. Recently, neural networks have emerged as the paramterization of choice for forward models \cite{loper2015smpl, pavlakos2019expressive, abdal2019image2stylegan, menon2020pulse, yu2021plenoxels, wang2019bidirectional, chen2021full, zhang2021unsupervised}, which can be pretrained on large collections of data, and inverted at testing time in order to solve inference queries.

\looseness=-1
Optimization-based inference has a number of advantages over feed-forward or encoder-based inference (EBI) for neural network inversion. Since there is no encoder, OBI provides flexibility to adapt to new tasks, allowing one to define new constraints into the objective during inference \cite{chen2021full,abdal2020image2stylegan}. When observations are partially missing, OBI can adapt without additional training \cite{menon2020pulse}. Moreover, OBI naturally supports generating multiple and diverse hypotheses when there is uncertainty \cite{shadows,menon2020pulse}. Finally, OBI has intrinsic advantages for robustness, both adapting to new data distributions as well as defending against adversarial examples \cite{mao2021adversarial}. 

However, the key bottleneck for OBI in practice is the computational efficiency and the speed of inference. Feedforward models are fast because they only require a single forward pass of a neural network, but OBI requires many (often hundreds) steps  of optimization in order to obtain strong results for one example. Forward models in OBI are often trained with generative or discriminative tasks, but they are never trained for the purpose of performing gradient descent in the input space. Consequently, the loss landscape is often highly non-convex, visualized in Fig.~\ref{fig:teaser1} (left). This non-convexity directly causes the instability and inefficiency of the optimization. 

In this paper, we propose a new framework which allows for faster and stabler optimization for OBI. Our key insight is to train a mapping network in the latent space that is aware of the optimization procedure at inference time. Existing optimization over the input space converges slowly, because the forward model is not designed for a test-time optimization procedure. By first collecting samples from the optimization trajectories in the latent input space, and then training the mapping network to minimize the loss on each sample of the trajectories, the mapping network will map each sample, even on the beginning of the trajectory, to an input that minimizes the loss. In doing this, the landscape becomes smoother, which allows even a few steps of optimization to obtain high quality results.
We achieve this with an coordinate descent algorithm that first collects samples from the optimization trajectories in the new input space, then trains the mapping network to minimize the loss on each sample of the trajectories.

Empirical experiments and visualizations on both generative and discriminative models show that our method can significantly improve the convergence speed for optimization. We validate our approach on a diverse set of computer vision tasks, where we achieve up to 34\% gain on GAN inversion~\cite{abdal2019image2stylegan}, 18\% accuracy gain on adversarial defense~\cite{mao2021adversarial}, and up to 75\% performance gain on 3D human pose reconstruction~\cite{pavlakos2019expressive}. In addition, our method converges an order of magnitude faster without loss in absolute performance. As our approach does not require retraining of the forward model, it can be compatible to all existing OBI methods with a differentiable forward model and objective function. 

The primary contribution of this paper is an efficient optimization-based inference framework. In Sec.~\ref{related_work}, we survey the related literature to provide an overview of forward model inversion problem. In Sec.~\ref{method}, we formally define OBI, our method to learn a faster loss landscape for OBI, and a training algorithm for better generalization and robustness. In Sec.~\ref{experiments}, we experimentally study and analyze the effectiveness of mapping network for OBI. We will release all code and models.

\section{Related Work} \label{related_work}

The different approaches for inference with a neural network can be partitioned into either encoder-based inference, which is feedforward, or optimization-based inference, which is iterative. We briefly review these two approaches in the context of our work.

% Existing work study how to infer a point estimate in the latent space that produce the target output. There are two major ways, encoder based method and backpropagation inference based method. Variational auto-encoder is an established way to infer the latent code for the generating the query sample. GAN inverse method training an encoder to infer the latent code for pretrained style-GAN.

\subsection{Encoder-based Inference}

Encoder-based inference trains a neural network $F$ to directly map from the output space 
to the input space. Auto-encoder based approach~\cite{pidhorskyi2020adversarial} learns an encoder that map the input data to the latent space. \cite{richardson2021encoding, tov2021designing, Wei2021ASB, perarnau2016invertible} learn an encoder from the image to the latent space in GAN. Encoder based inference requires training the encoder on the anticipated distribution in advance, which is often less effective and can fail on unexpected samples~\cite{dinh2021hyperinverter, kang2021gan}. In addition, encoder-based method can only produce one reconstruction for image inpainting~\cite{yeh2017semantic}, even if multiple outcomes can all be true given the partial input.

\subsection{Optimization-based Inference}

OBI methods perform inference by solving an optimization problem with gradient-based methods such as Stochastic Gradient Descent (SGD)~\cite{SGD} and Projected Gradient Descent (PGD)~\cite{madry2017towards}.
In these cases, the objective function specifies the inference task.
Besides these methods which use a point estimate for the latent variable, one can estimate the posterior distribution of the latent variable through Bayesian optimization, such as SGLD~\cite{welling2011bayesian}. 

Gradient based optimization methods have been used to infer the latent code of query samples in deep generative models like GANs~\cite{goodfellow2014generative} via GAN inversion \cite{karras2020analyzing, jahanian2019steerability, shen2020interpreting, zhu2016generative, abdal2019image2stylegan, Abdal_2020_CVPR, bau2019seeing, huh2020transforming, pan2021exploiting}. 
Style transfer relies on gradient based optimization to change the style of the input images~\cite{jing2019neural}.  It can also create adversarial attacks that fool the classifier~ \cite{AA, CW, TLA, intriguing}. Recently, backpropagation-based optimization has shown to be effective in defending adversarial examples~\cite{mao2021adversarial}. 

Recently, constrained optimization was popularized for text-to-image synthesis by \cite{Crowson2022VQGANCLIPOD, Liu2021FuseDreamTT}. They search in the latent space to produce an image that has the highest similarity with the given text as measured by a multi-modal similarity model like CLIP \cite{Radford2021LearningTV}. Test-time constrained optimization is also related to the idea of `prompt-tuning' for large language models. \cite{Lester2021ThePO} learn ``soft prompts'' to condition frozen language models to perform specific downstream tasks. Soft prompts are learned through backpropagation to incorporate signal from just a few labeled examples (few-shot learning).

% However, the success of gradient based method relies on the convexity and smoothness of the loss landscape. To allow better optimization and inference for the latent via gradient descent,  \blue{ }

\looseness=-1 A major challenge for optimization-based inference is how to perform efficient optimization in a highly non-convex space. To address this, input convex model~\cite{icnn} was proposed so that gradient descent can be performed in a convex space. \cite{tripp2020sample} introduced a method to retrain the generative model such as it learns a latent manifold that is easy to optimize. When the model cannot be changed and retrained, bidirectional inference~\cite{wang2019bidirectional} and hybrid inference~\cite{zhu2016generative, zhu2020domain} uses an encoder to provide a good initialization for the optimization-based inference in a non-convex space. Our method does not retrain the generative model, but maps the original latent space to a fast loss landscape.

% \subsection{Meta-Learning }

% \subsection{GAN Inversion}

\section{Learning  Landscapes for Fast Inference} \label{method}

\begin{figure}[t!]
    \centering
    \includegraphics[width=\linewidth]{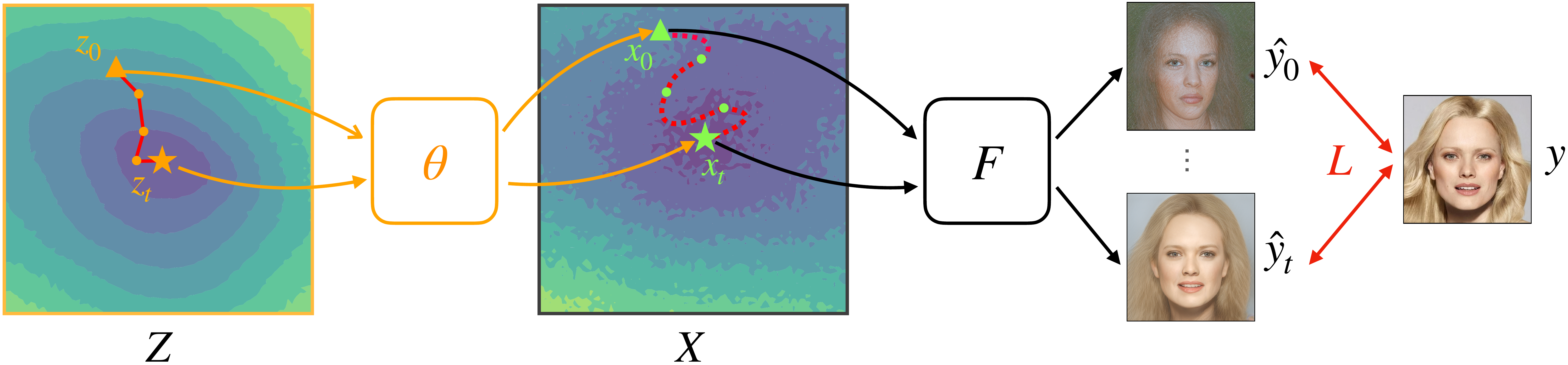}
    \caption{\textbf{Method}. The left and middle figure show the loss landscape for our latent space and the original latent space, respectively. While walking to the optimal solution in a few steps is hard in $X$ space, it can be done in our learned loss landscapes.}
    \label{fig:method}
    \vspace{-1em}
\end{figure}

We present our framework to learn a fast loss landscape for optimization-based inference (OBI) methods. In Sec.~\ref{obi_def}, we will define OBI. In Sec.~\ref{objective}, we will introduce our framework as well as the training objective. In Sec.~\ref{training}, we will describe how to train our model with a coordinate descent algorithm and an experience-replay buffer.

\subsection{Optimization-based Inference} \label{obi_def}

Let $F(\x) = \hat{y}$ be a differentiable forward model that generates an output $\hat{y}$ given an input variable $\x \in X$. For example, $\hat{y}$ might be an image, and $\x$ might be the latent variables for a generative model. Given an observation $y$, the goal of OBI is to find the input $\hat{\x} \in X$ such that an objective function $L(\hat{y}, y)$ is minimized. Formally, we write this procedure as:
\begin{equation}
    \hat{\x} = \argmin_{\x \in X} L(F(\x), y)
\end{equation}
When the objective function L and the model $F$ are both differentiable, we can perform the optimization over input space $X$ with gradient descent. However, the original forward model has not been trained to perform gradient descent in the input space. Consequently, the loss landscape is highly non-convex, making the convergence slow.

\subsection{Remapping the Input Space} \label{objective}
Instead of operating in the original input space $X$, we will create a new space $Z$ where gradient descent is efficient and converges in a small number of iterations. To parameterize $Z$, we will use a neural network $\theta : Z \rightarrow X$ that maps from the new space $Z$ to the original space $X$. The learning problem we need solve is to estimate the parameters of $\theta$ so that there is a short gradient descent path in $Z$ from the  initialization to the solution. Fig.~\ref{fig:method} shows an overview of this  setup.

\begin{SCfigure}
    \centering
    \includegraphics[width=0.68 \linewidth]{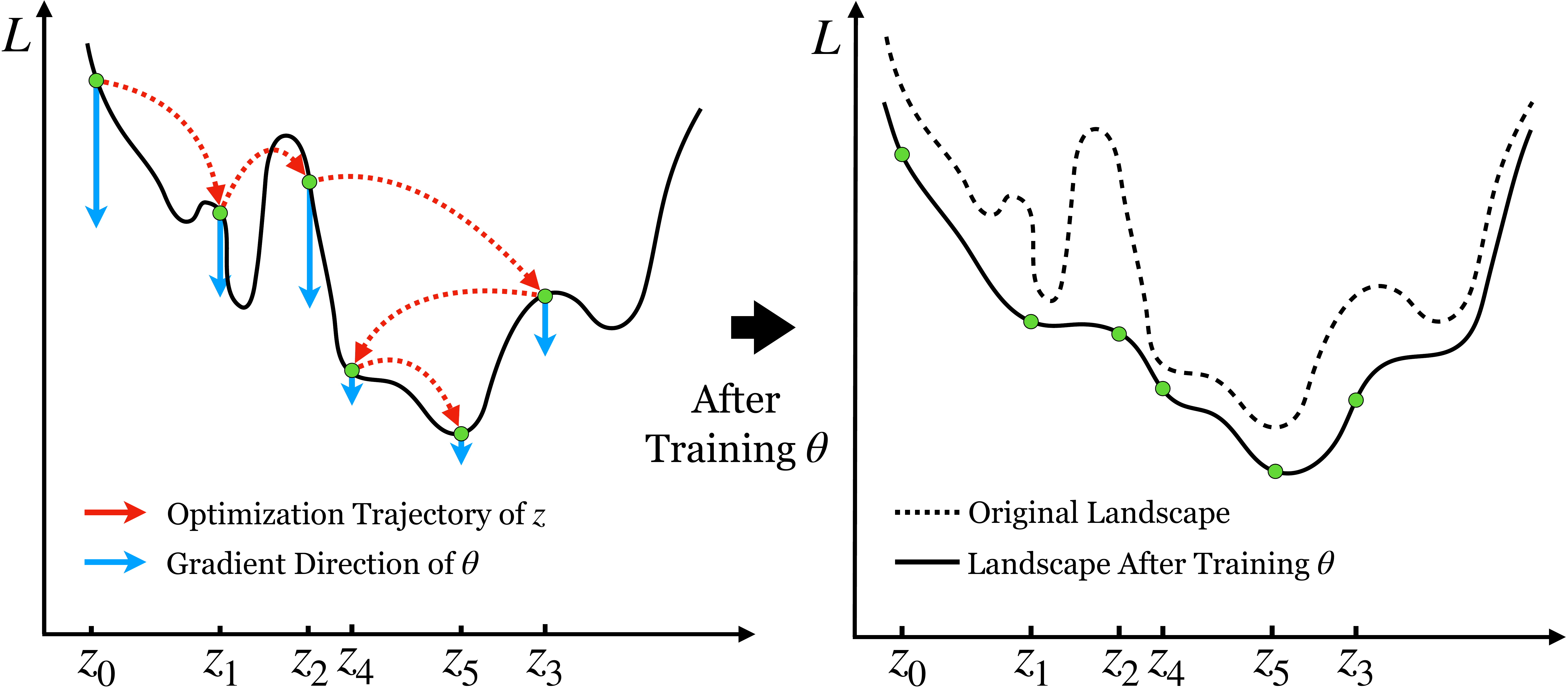}
    \caption{\textbf{Landscape Learning.} An optimization trajectory $\{\z_t\}_{t=0}^{5}$ collected is used to train $\theta$. $\z_i$ that corresponds to a higher $L_i$ will yield a higher gradient when training $\theta$. Optimization over multiple steps along the trajectory causes $\theta$ to learn \textit{patterns} of trajectories and create a smoother loss landscape.}
    \label{fig:intuition}
\end{SCfigure}

We formulate an objective by rolling out the gradient updates on $\z$, where we write $\z_t \leftarrow \z_{t-1} + \lambda \frac{\partial L}{\partial \z_{t-1}}$ as the $t^{\text{th}}$ update with a step size of $
\lambda$. For gradient descent in space $Z$ to be efficient, the goal of our introduced $\theta$ is to move every step $\z_t$ as close as possible to the global minima:
% \begin{equation}
%     \hat{\theta}= \argmin_{\theta} \sum_i \min_{\z_i} L(F(\theta(\z_i)), y_i) \label{eq:overall_objective}
% \end{equation}
\begin{equation}
    \hat{\theta}= \argmin_{\theta} \; \mathbb{E}_{(\z, y)} \left[ \sum_{t=0}^T L(F(\theta(\z_{t})), y) \right] \quad \textrm{where} \quad \z_t = \begin{cases}  0, & t = 0\\
    \z_{t-1} + \lambda \frac{\partial L}{\partial \z_{t-1}}, & t > 0 \end{cases} \label{eq:overall_objective}
\end{equation}
We visualize this process with a toy example in Fig.~\ref{fig:intuition}. Gradient updates on $\theta$ w.r.t multiple steps $\z_t$ along a trajectory will cause the loss value on each step to be lowered.  By learning the parameters of $\theta$ across many examples, $\theta$ can learn the patterns of optimization trajectories in $X$. For example, $\theta$ can learn to estimate the step size in $X$ and dynamically adjust it to each example. Moreover, $\theta$ can learn to smooth non-convex regions in the landscape. 

Once we obtain $\hat{\theta}$, we do inference on a new example $y$ through the standard optimization-based inference procedure, except in $Z$ space now. Given the observation $y$, we find the corresponding $\hat{\x}$ through the optimization:
\begin{equation}
    \hat{\x} = \hat{\theta}(\hat{\z}) \quad \textrm{where} \quad 
    \hat{\z}= \argmin_{\z \in Z} \; L(F(\hat{\theta}(\z)), y)
\end{equation}
When the inverse problem is under-constrained, one can infer multiple hypotheses for $\hat{\x}$ by repeating the above optimization multiple times with a different random initialization for $\z_0$.

\vspace{-0.5em}

\subsection{Training} \label{training}

% We train $\theta$ in Equation (2) with a hard EM algorithm \cite{mcallester2007tticem}. First we rewrite the objective Eq. \ref{eq:overall_objective} as:
% \begin{equation}
%     \hat{\theta}= \argmax_{\theta} \prod_i P_\theta(y_i|F(\theta(\z_i))) \label{eq:mle_objective}
% \end{equation}
% To optimize objective Eq. \ref{eq:mle_objective}, we perform the following algorithm:
% \begin{enumerate}
%     \item Initialize $\theta$ with Gaussian distribution
%     \item Repeat the following until $\prod_i \ln P_\theta(y_i|F(\theta(\z_i)))$ converges:
%     \begin{enumerate}
%         \item \textit{Expectation}: $\rho_\theta(z_i) = \delta(z_i = \tilde{z}_i)$, where $\tilde{z}_i = \argmax_{z_i} P_\theta(y_i|F(\theta(\z_i))))$
%         \item \textit{Maximization}: $\hat{\theta} = \argmax_{\theta} \mathbb{E}_{z\sim \rho}[\ln(P_\theta(y_i, F(\theta(\z_i))))]$
%     \end{enumerate}
% \end{enumerate}

% Here $\delta(\cdot)$ denotes a Dirac delta distribution. During the expectation stage, we fix the weights in $\theta$ and optimize $\hat{z}$ with gradient descent to minimize the objective function Eq. \ref{eq:overall_objective}. Then during the maximization stage, we fix the previously collected $\hat{z}_i$, where $N$ is the number of optimization steps for each trajectory, to optimize $\theta$ with another pass through the entire forward and backward propagation process.

We use coordinate descent (CD) in order to train $\theta$ jointly with estimating $\z$ for each example in the training set. Specifically, we  first fix parameters of $\theta$ and collect $N$ optimization trajectories of $\z$, each with length $T$. Adopting the same terminology from the robotics community for learning on continuous states \cite{mnih2013playing}, we term this a \textit{experience replay buffer}. Subsequently, we randomly sample data from this buffer and train $\theta$ to optimize the loss function. We alternate between these two steps for a fixed number of iterations with gradient descent for both. Depending on the application, the training time for $\theta$ varied from one hour to one day using a four GPU server.  
Please see the appendix for more implementation details.

We also experimented with an online version of the training algorithm, where we update $\theta$ immediately after one update to $\z$. However, in our experiments, we found this resulted in a slower convergence rate. We show these comparisons in the ablation experiments. 

\section{Experiments} \label{experiments}

\begin{figure}%
    \centering
    \subfloat[\centering GAN Inversion]{{\includegraphics[width=6.3cm]{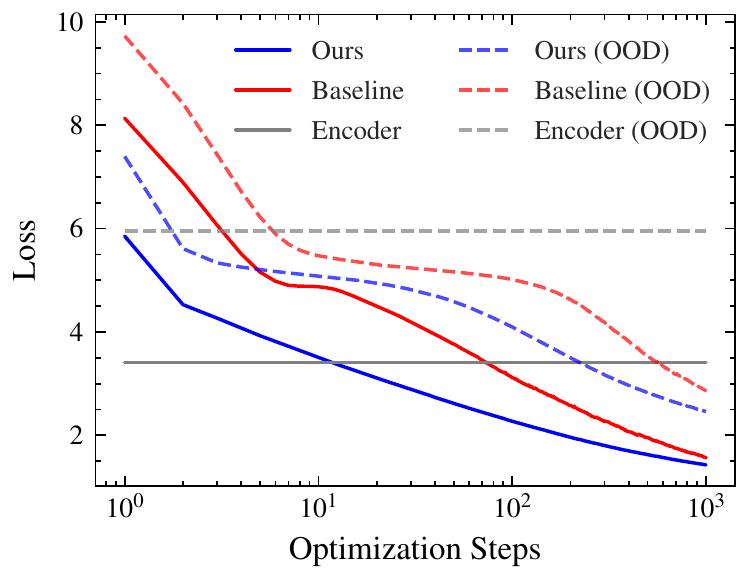} } \label{fig:quant_gan}}%
    \subfloat[\centering 3D Human Pose Reconstruction]{{\includegraphics[width=6.5cm]{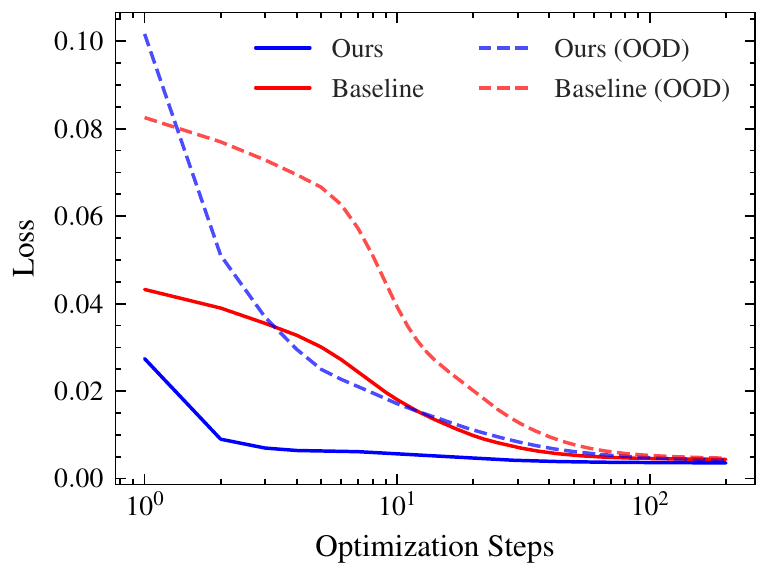} }\label{fig:quant_3dhuman}}%
    \caption{\textbf{Optimization Performance.} \looseness=-1 We visualize the trends of optimization performance compared with the baseline. In \textbf{GAN Inversion (Left)}, we evaluate all models on test splits of CelebA-HQ \cite{karras2017progressive} and LSUN-cat \cite{yu2015lsun} (OOD) with loss defined in Eq. \ref{eq:loss_gan}. 
    %The encoder model \cite{richardson2021encoding} was trained on CelebA-HQ training split. 
    Since encoder-based inference doesn't involve optimization, we use a flat line to represent it. We perform 2000 steps of gradient descent for all models except encoder-based models. In \textbf{3D Human Pose Reconstruction (Right)}, we evaluate all models on test splits of GRAB \cite{taheri2020grab} and PROX \cite{hassan2019resolving} (OOD) with loss defined in Eq. \ref{eq:loss_human}. We perform 200 steps of gradient descent for all models. 
    For each step, we plot the average loss value of test splits.}%
    \label{fig:quantitative}%
    \vspace{-0.5em}
\end{figure}

The goal of our experiments is to analyze how well our proposed method can be applied to various existing OBI methods to improve the optimization efficiency. We demonstrate application of our method to three diverse OBI methods in computer vision, including both generative models and discriminative models. For each OBI method, the inference-time optimization objective of the baseline and ours can be written as:
\begin{equation} \label{eq:inference_obj}
    \text{Baseline:  } \hat{\x} = \min_{\x \in X} L(F(\x), y), \hspace{0.2in} \text{Ours:  } \hat{\z} = \min_{\z \in Z} L(F(\theta(\z)), y)
\end{equation}
Next, we provide the specific implementation of the loss term $L$ for each application, along with quantitative and qualitative results. We also perform experiments to understand the loss landscape in Sec.~\ref{loss_landscape} and perform ablations on different parts of our approach in Sec.~\ref{ablations}.

% ----------
\subsection{GAN Inversion}
\looseness=-1
 We first validate our method on StyleGAN inversion~\cite{abdal2019image2stylegan}. We take a pretrained generator of StyleGAN \cite{karras2019style} denoted as $F$. Let $y$ be an observed image whose input variable we are recovering, we optimize the objective of Eq. \ref{eq:inference_obj} where the loss can be written as,
\begin{equation}
    L(\hat{y}, y) = L_{lpips}(\hat{y}, y) + || \hat{y} - y ||_2^2 \label{eq:loss_gan}
\end{equation}
where $L_{lpips}$ is a perceptual similarity loss introduced in \cite{zhang2018unreasonable}, $\hat{y} = F(\hat{x})$ for baseline and $\hat{y} = F(\theta(\hat{z}))$ for ours. We train $\theta$ on the train split of CelebA-HQ\cite{karras2017progressive} dataset and evaluate on CelebA-HQ validation split for in-distribution experiments and LSUN-Cat\cite{yu2015lsun} for distribution shifting (OOD) experiments. We compare the results from our method against the state-of-the-art encoder-based GAN inversion model \cite{richardson2021encoding}.

\textbf{Quantitative Results.} From Fig.~\ref{fig:quant_gan}, we see that in all experiments, optimization in our space $Z$ consistently outperforms the baseline from the first optimization step to after convergence. This gap in performance is even larger when evaluated on OOD data. This suggests that the improvement in performance is not caused by memorizing the training data. Note that our image reconstruction performance after only 2 steps of optimization is able to outperform or be on-par with the baseline at 20 steps of optimization, resulting in a 10-fold improvement in efficiency. Even after convergence (after 2000 optimization steps), our reconstruction performance improves over the baseline by 15\% for in-distribution evaluation and 10\% for OOD evaluation. When compared with encoder-based GAN inversion \cite{richardson2021encoding}, our method achieves better reconstruction after 11 steps of optimization for in-distribution data and 3 steps for OOD data. 

\begin{figure}[t!]
    \centering
    \includegraphics[width=\linewidth]{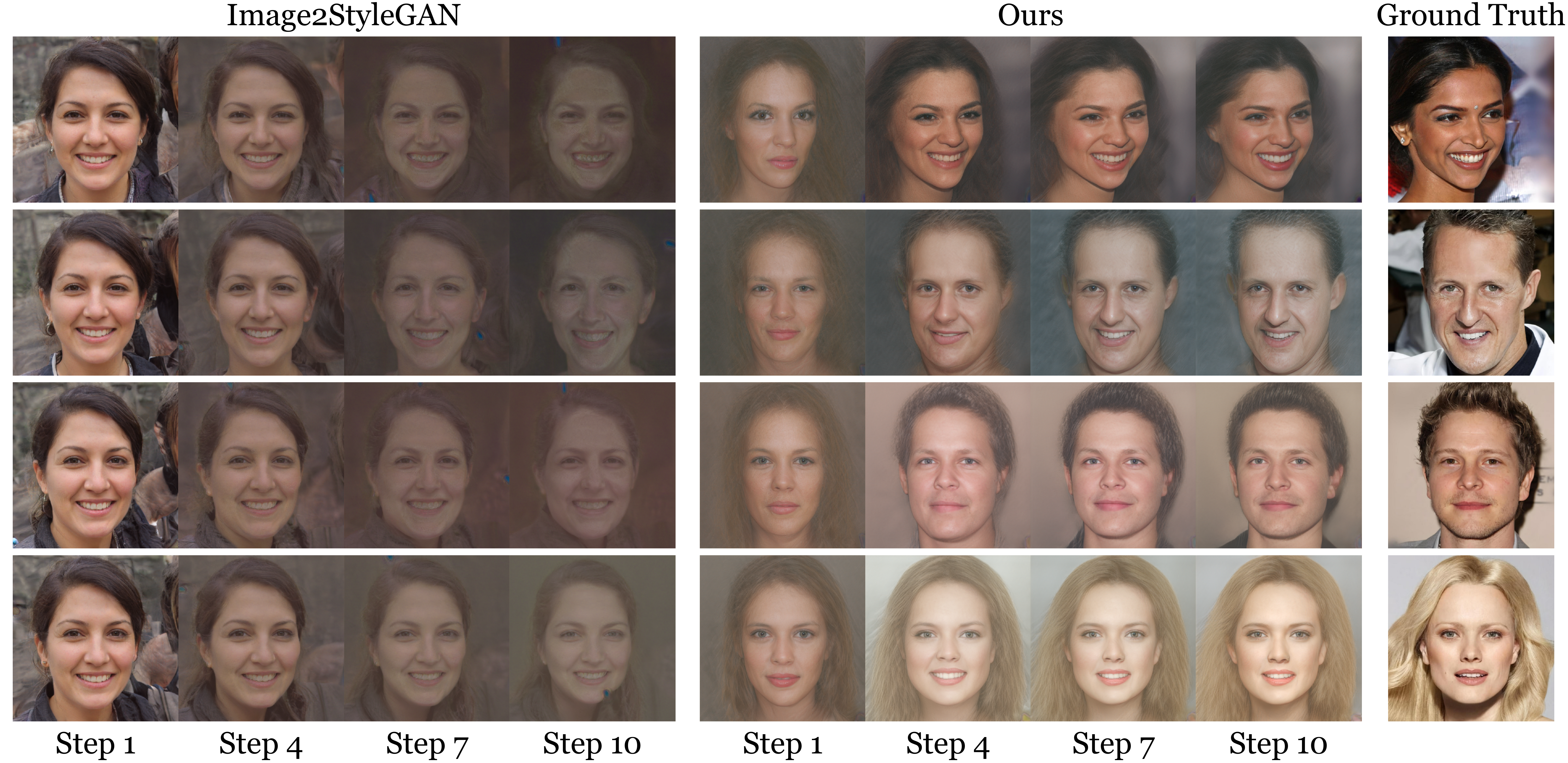}
    \caption{\textbf{Optimization Process for GAN Inversion.}  Comparing optimization process of our method and the baseline in order to reconstruct the ground truth image. \textbf{Left} shows the results from the baseline where optimization is done in the original input space $X$. \textbf{Middle} shows the results from our method where optimization is done in our space $Z$. \textbf{Right} column contains the ground truth image to each example. Each row corresponds to the same example.} 
    \label{fig:optimization}
    \vspace{-1em}
\end{figure}

\begin{figure}[t!]
    \centering
    \vspace{-0.3em}
    \includegraphics[width=0.9\linewidth]{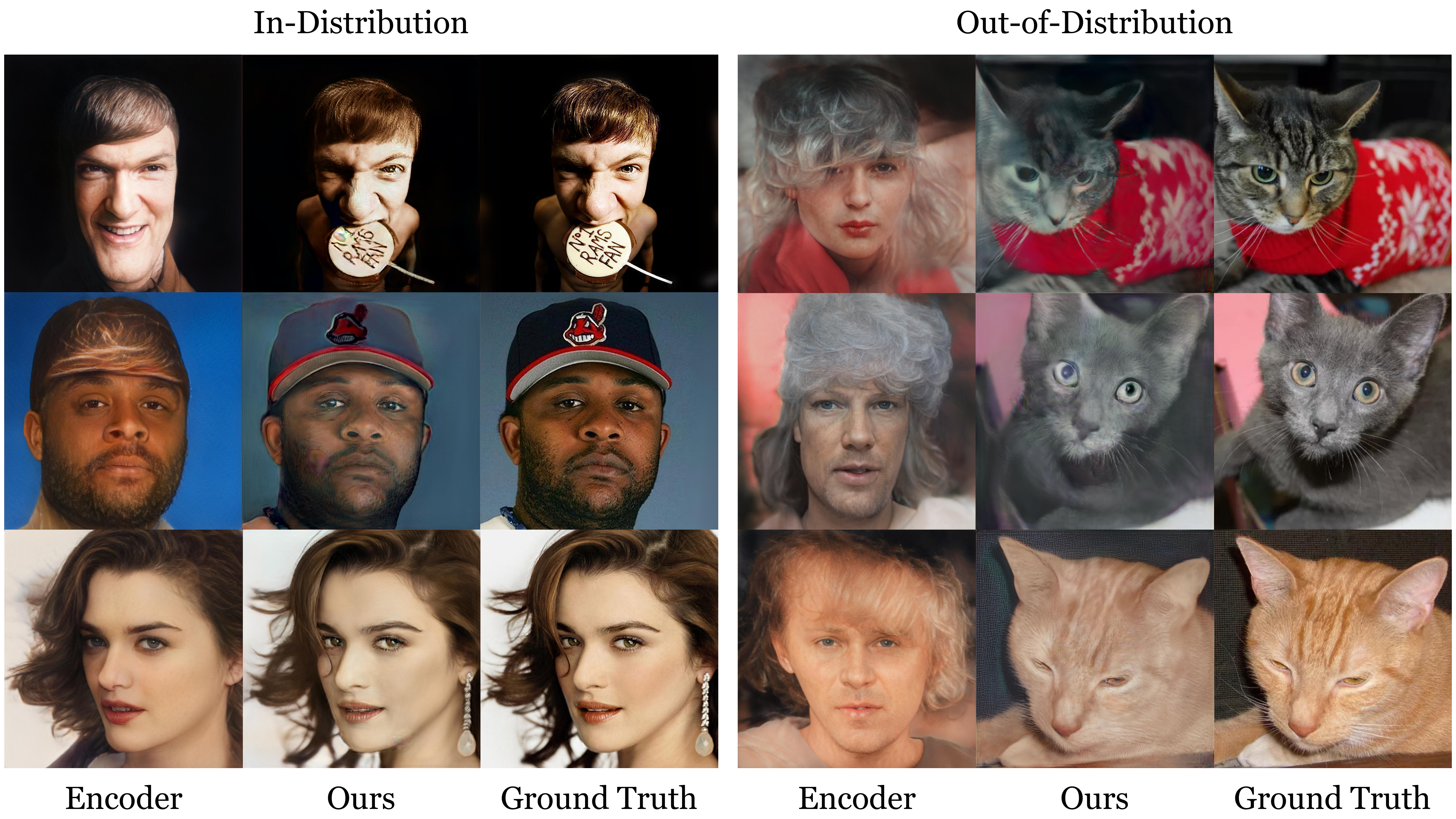}
    \caption{\textbf{Comparison Against Encoder-Based Inference.} \textbf{Left} shows the results on the test split of CelebA-HQ; \textbf{Right} shows the results on the LSUN-cat dataset.}
    \label{fig:encoder}
    \vspace{-0.7em}
\end{figure}

\textbf{Qualitative Results.} From Fig.~\ref{fig:encoder}, we can see that our method already shows improvements on in-distribution data - it can almost perfectly reconstruct details like fine hair strands, the cap on the person's head, the object in the person's mouth as well-as the text on it. 
Interestingly, our method is able to discover and reconstruct latents for cats while the encoder-based model fails miserably as shown in Fig~\ref{fig:encoder}. 
The performance on OOD data truly highlights the benefits of our method.
We also visualize how the face generations evolve over the process of optimization in Fig.~\ref{fig:optimization}. 
We can see that in just 4 steps, our method is already able to reconstruct coarse features such as face orientation, skin tone and hair color, while the baseline has hardly deviated from the initialization in regard to any of these features. 
Further, in Fig.~\ref{fig:diversity} we visualize reconstructions from partial observations where only the center of the face (row 1) or everything other than the mouth (row 2) is visible. We can see a variety of feasible possibilities for the hidden regions (e.g., different hairstyles, lip colors, expressions, etc) showcasing the diversity of the new latent space.

% ----------

\begin{figure}[t!]
    \centering
    \includegraphics[trim={1cm 4.5cm 1cm 0.5cm},clip,width=0.97\linewidth]{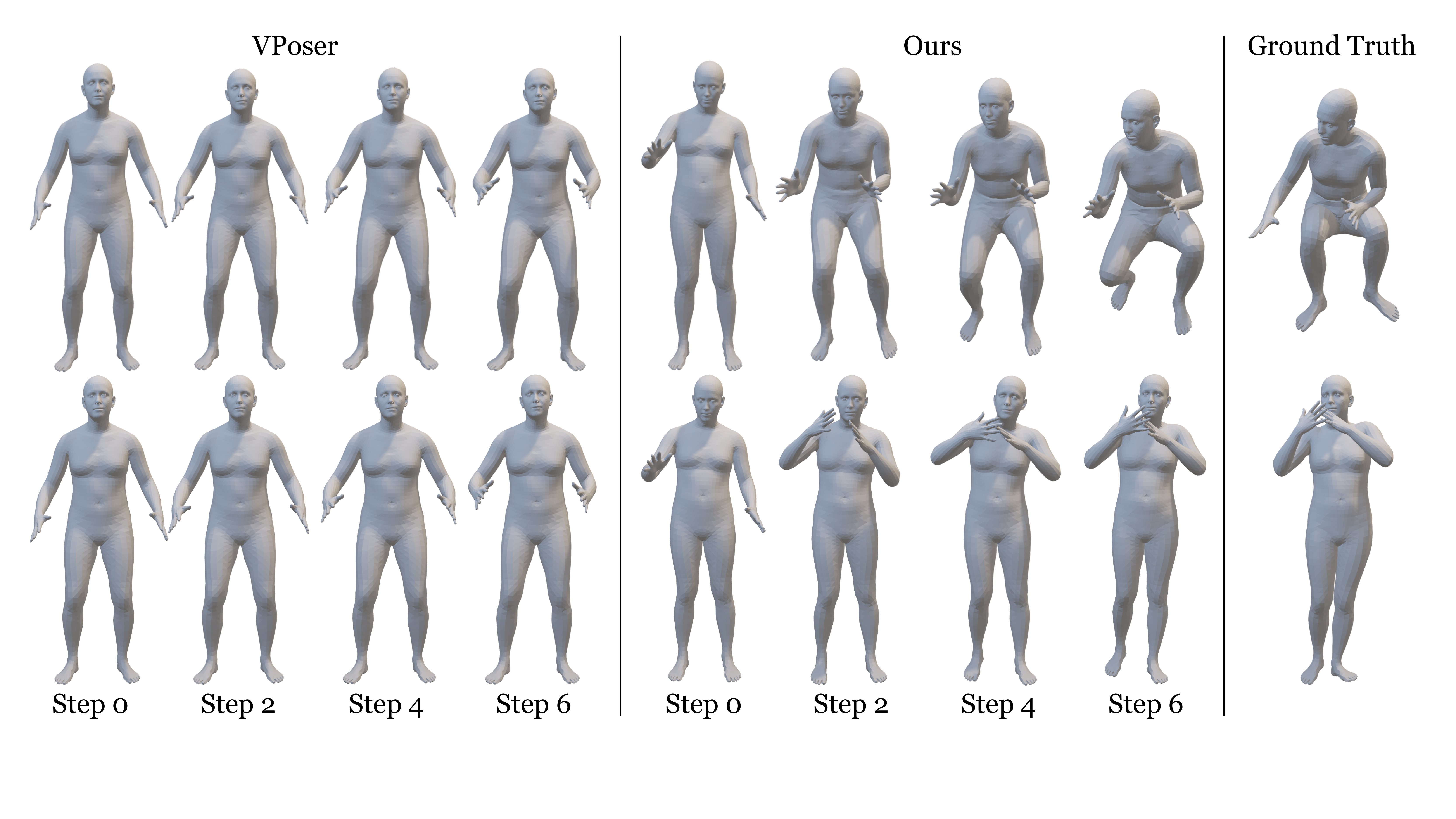}
    \caption{\textbf{Optimization Process for 3D Human Pose Reconstruction.} Results shown are for out-of-distribution PROX dataset for sitting (\textbf{Top}) and standing (\textbf{Bottom}) poses.} 
    \label{fig:human_opt}
    \vspace{-1em}
\end{figure}

\subsection{3D Human Pose Reconstruction}
In addition to image generation, our framework also works for 3D reconstruction. For this, we use VPoser \cite{pavlakos2019expressive} -- a variational autoencoder \cite{kingma2013auto} that has learnt a pose prior over body pose. 
VPoser was trained on SMPL-X body pose parameters $y \in \mathbb{R}^{63}$ obtained by applying MoSh \cite{mosh} on three publicly available human motion capture datasets: CMU \cite{cmumocap}, training set of Human3.6M \cite{human36m}, and the PosePrior dataset \cite{poseprior}. 

We take a pretrained VPoser decoder denoted as $F$.  Our trained mapping network $\theta$ projects a vector from the new input space $\z \in Z$ to a vector in the original VPoser decoder's input space $\x \in X$. 
Similar to GAN Inversion, we optimize the objective of Eq. \ref{eq:inference_obj}, where the loss function between predicted and ground truth pose parameters is,
\begin{equation}
    L(\hat{y}, y) = || \hat{y} - y ||_2^2  \label{eq:loss_human}
\end{equation}
where $\hat{y}=F(\hat{x})$ for the baseline and $\hat{y} = F(\theta(\hat{z}))$ for ours.
For training $\theta$, we use the GRAB dataset~\cite{taheri2020grab} which contains poses of humans interacting with everyday objects.
We construct splits for novel video sequences -- thus the test split will contain a seen human subject but a potentially unseen pose / demonstration by that subject. 
We evaluate on this test split for in-distribution experiments and on the PROX dataset~\cite{hassan2019resolving} for OOD experiments, which contains poses of humans interacting in 3D scenes (e.g., living room, office, etc).

\textbf{Quantitative Results.} For SMPL-X human pose reconstruction experiment, the results follow a similar trend as GAN inversion, with massive improvement in both convergence speed and final loss values after convergence (see Fig.~\ref{fig:quant_3dhuman}). Our method outperforms the baseline by 19\% for in-distribution evaluation and 11\% for OOD evaluation.

\begin{figure}[t!]
    \centering
    \includegraphics[width=0.9\linewidth]{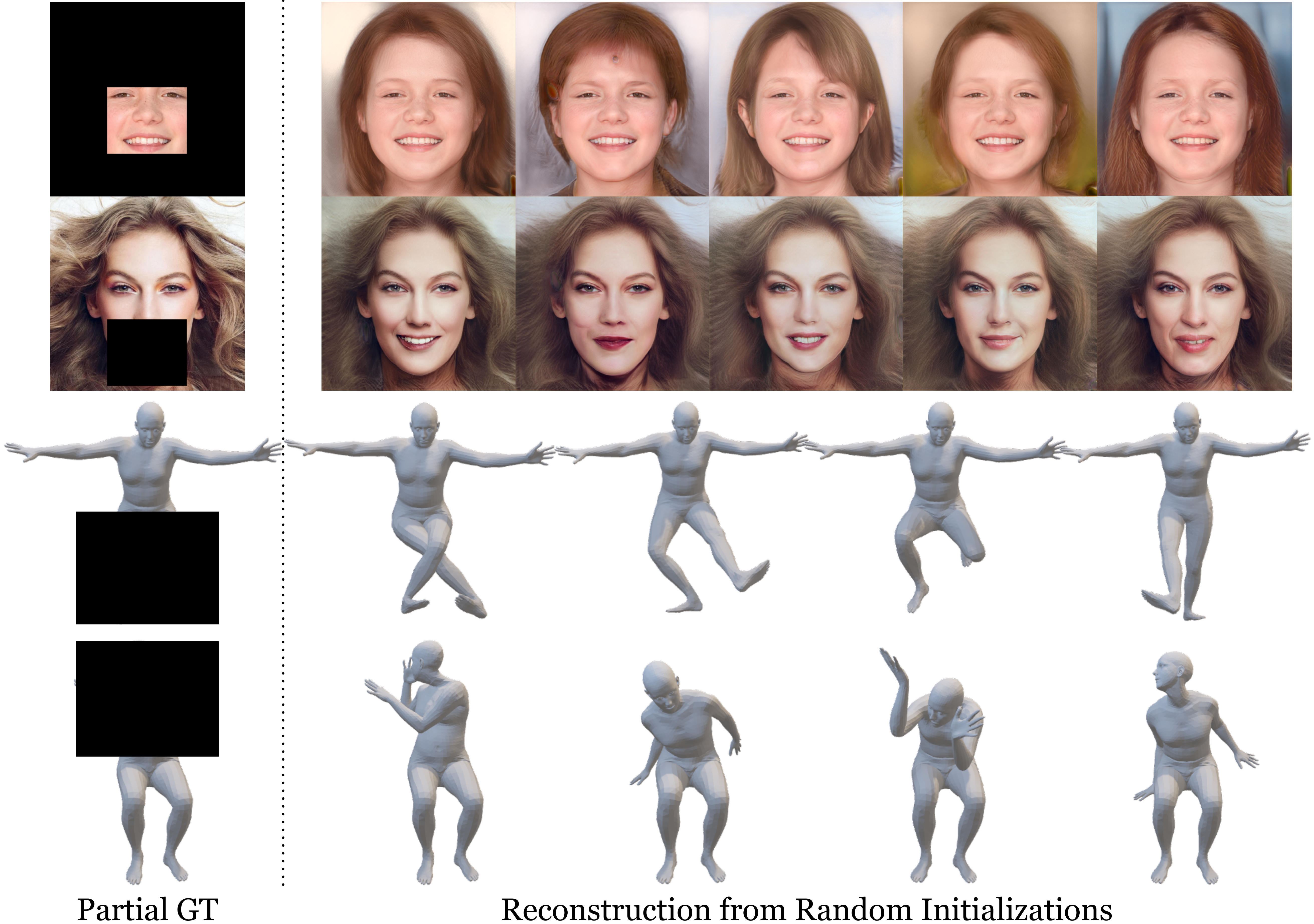}
    % \vspace{-0.5em}
    \caption{\textbf{Diversity of Masked Reconstructions.} We visualize reconstructions for partially observable inputs from random initializations. The masked regions are not considered for loss computation, i.e., the gradient is set to be zero. By optimizing only on the partial observation, we obtain diverse, feasible solutions for the hidden regions.}
    \label{fig:diversity}
    \vspace{-1em}
\end{figure}

\textbf{Qualitative Results.} In Fig.~\ref{fig:human_opt} we visualize how the human pose reconstructions evolve over the process of optimization. Here, we observe that the reconstructions from steps 0 to 6 of the baseline are similar for both examples. On the other hand, our method caters to fast convergence for the varying examples, highlighting the general, yet efficient properties of our search space. Further, in Fig.~\ref{fig:diversity} we visualize reconstructions from partial observations where the only joints visible are that of the upper body (row 3) or lower body (row 4). We obtain a wide range of feasible possibilities for the hidden joints demonstrating the diversity of the latent space.
\vspace{-1em}
% ----------
\subsection{Defending Adversarial Attacks}

Our method is also applicable to discriminative models. A state-of-the-art defense~\cite{mao2021adversarial} for adversarial attack optimizes the self-supervision task at inference time, such that the model can dynamically restore the corrupted structure from the input image for robust inference. Following the existing algorithm implementation, we optimize the input image via our method by minimizing the following discriminative loss function: 
% \begin{equation}
%   \r = \argmin_{\r} \mathcal{L}_s(\x_a + \r)  = \argmin_{\r}\mathbb{E}_{i,j}\left[-\y_{ij}^{(s)} 
%     \log \frac{\exp(\mathrm{cos}(\f_i, \f_j)/\tau)}{\sum_{k}\exp(\mathrm{cos}(\f_i, \f_k)/\tau)}  
%     \right],
%      \label{eq:loss_adv}
% \end{equation}
\begin{equation}
   L(F(\r + \a), y) = L(F(\theta(\z) + \a), y) = \mathbb{E}_{i,j}\left[-y_{ij}^{(s)} 
    \log \frac{\exp(\mathrm{cos}(\f_i, \f_j))}{\sum_{k}\exp(\mathrm{cos}(\f_i, \f_k))}  
    \right] + \lambda ||\theta(\z)||_2^2,
     \label{eq:loss_adv}
\end{equation}
%+ \lambda ||\r||_2^2
where $\a$ is the adversarial attacks that we aim to defend against, $\r=\theta(\z)$ is our additive defense vector to optimize, $\f_i$ are the contrastive features produced by the neural network $F$ from the $i^{\text{th}}$ image instance, and $\y_{ij}^{(s)}$ is the indicator for the positive pairs and negative pairs. 

After obtaining the mapping network $\theta$ and the input variable $\hat{\z}$, the prediction is $\hat{\y} =  F'(\a+\theta(\z))$, where $F'$ is the classification model. Note that the a self-supervision loss is optimized as a proxy for increasing the robust classification accuracy. In addition, we add a $L_2$ norm decay term for the generated noise $z$ to avoid generating reversal vector that is too large. 

\textbf{Quantitative Results.} We evaluate our method on four popular pretrained robust models~\cite{rice2020overfitting, MART, AWP, unlabeled} on CIFAR-10 dataset. The results in \cite{mao2021adversarial} require many steps to optimize the objective to improve the adversarial robustness, which slows down the inference by hundreds of times than the original forward pass. Ideally, we desire test-time optimization that can adapt to the attacked images in just one step, causing the minimal delay at inference time. In Table~\ref{tab:attack}, our method outperforms the gradient descent method in \cite{mao2021adversarial} by up to 18\% robust accuracy using a single step, providing a real-time robust inference method. Note that our method converges after 1 step of optimization, demonstrating the effectiveness of our approach.
\begin{table}[]
\centering
\begin{tabular}{l|c|cc|cc|cc}
\toprule
% \multicolumn{8}{c}{\textbf{Adversarial Attack (results not entered yet)}} \\ \midrule
&  \multicolumn{7}{c}{Optimization Steps} \\
 & No & \multicolumn{2}{c}{1 step}    & \multicolumn{2}{c}{3 steps}    & \multicolumn{2}{c}{5 steps}  \\ 
  Model Type  & Optimization  & Baseline & Ours   & Baseline & Ours   & Baseline & Ours  \\ 
\midrule
RO~\cite{rice2020overfitting} & 31.99 & 34.62 & \textbf{44.65} & 36.77 & \textbf{44.23} & 38.38 & \textbf{43.43}\\
AWP \cite{AWP} & 35.61 & 39.54 & \textbf{51.39} & 42.81 & \textbf{51.67} & 44.96 & \textbf{51.05} \\
MART \cite{MART} & 35.66 & 39.77 & \textbf{51.77}& 42.50 & \textbf{51.77} & 45.42 & \textbf{50.96}  \\
SemiSL \cite{unlabeled} & 29.78 & 34.53 & \textbf{52.11} & 37.27 & \textbf{51.23} & 40.93 & \textbf{49.83}    \\
\bottomrule
\end{tabular}
\vspace{0.5em}
\caption{Experiment on improving adversarial robust accuracy. Our goal is to defend 200 steps of $L_2=256/255$ norm bounded attack~\cite{madry2017towards}, where the attack's step size is $64/255$. Our baseline is the SOTA test-time optimization-based defense~\cite{mao2021adversarial}, which minimizes the loss of self-supervision task.}
%Our approach can improve robust accuracy more effectively than the state-of-the-art by 18\%, achieving high robust accuracy with just one step of optimization.
\label{tab:attack}%
\vspace{-1em}
\end{table}
\vspace{-0.6em}
\subsection{Loss Landscape} \label{loss_landscape}
\begin{figure}[t!]
    \centering
    \vspace{-0.2em}
    \includegraphics[trim={1.3cm 5.8cm 1.3cm 5.8cm},clip,width=0.9\linewidth]{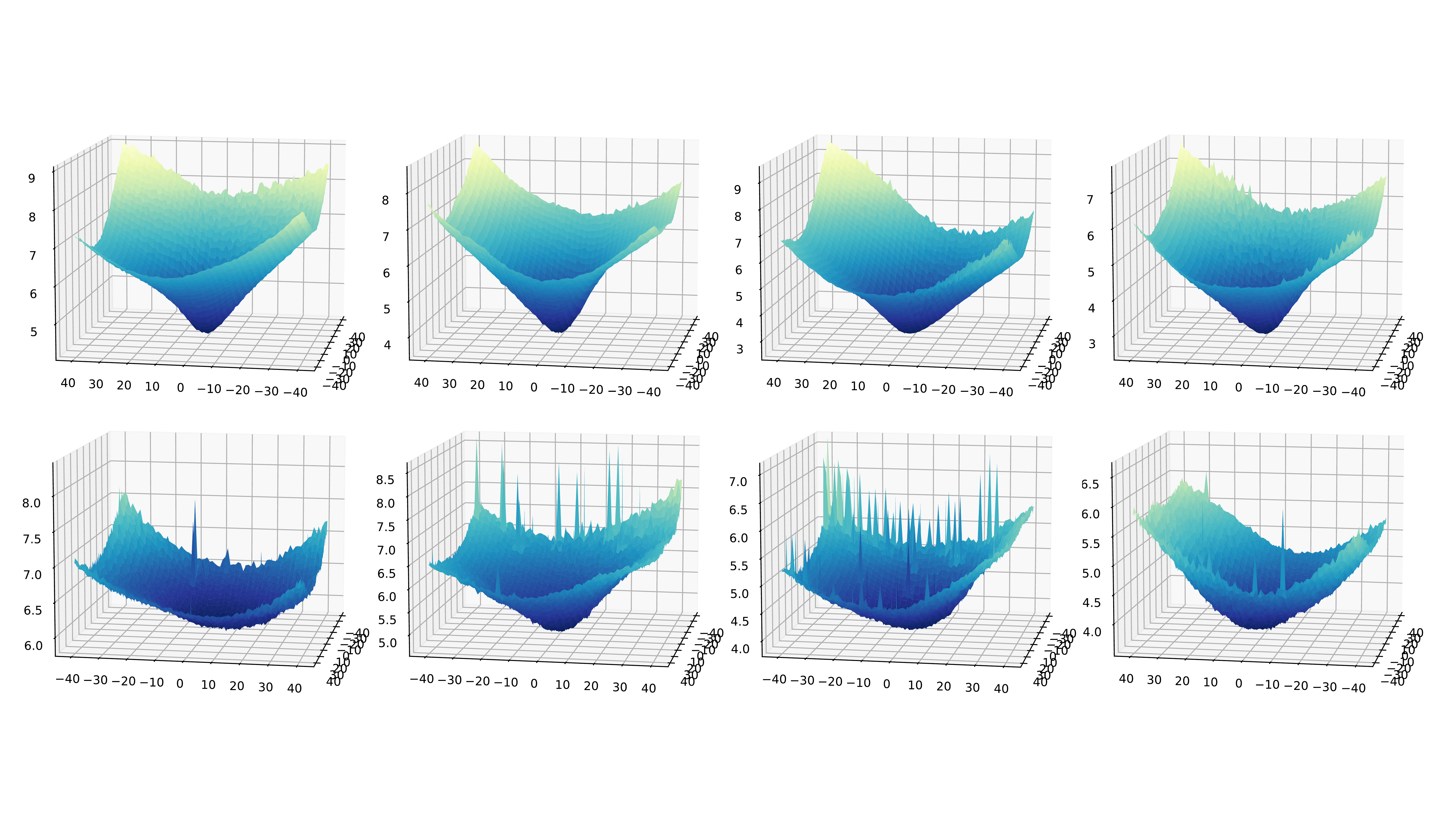}
    \caption{\textbf{Visualizing Loss Landscape (Uncurated).} Visualizing the loss landscape of StyleGAN inversion spanned by two principle directions. \textbf{Top} row shows 4 examples of the loss landscapes corresponding to our space $Z$. \textbf{Bottom} row shows the loss landscapes corresponding to the original input space $X$ for the same 4 examples.}  
    \label{fig:loss_landscape}
    \vspace{-0.2em}
\end{figure}

\vspace{-0.2em}
To understand the underlying cause of the significant improvement in optimization brought by our mapping network $\theta$, we visualize in Fig.~\ref{fig:loss_landscape} the loss landscape for performing optimization in the original input space $X$, and our projected space $Z$. To generate this visualization, we first perform 20 steps of optimization on the validation dataset to collect a set of recovered latents. We then perform principle component analysis (PCA) on these recovered latents to obtain two principle directions. Finally, for individual examples, we evaluate the loss for vertices on a meshgrid spanned by the two principle directions.

From the visualization, we can see that the loss landscapes of the baseline are highly non-convex and contain points whose loss are significantly higher than its neighboring regions, while our loss landscapes are significantly smoother, with the ``spikes'' removed. Besides, our loss landscapes also tend to be steeper than the baseline ones. These two phenomena directly cause our method to perform gradient descent faster and stabler.

\vspace{-0.3em}
\subsection{Ablation Study} \label{ablations}
\vspace{-0.2em}
In this section, we present an ablation study by removing the proposed coordinate descent scheme and the experience replay buffer. We also compare against a $\theta$ that is randomly initialized. From Table~\ref{tab:ablation}, we discovered that for in-distribution, coordinate descent and training of theta improves the optimziation performance by 14\% and 30\% respectively. Such gap becomes 35\% and 28\% for evaluation on 200 steps. For OOD data, the advantage is furthered enlarged as shown in Table~\ref{tab:ablation}.

One surprising result we discovered experimentally is that OBI under a randomly initialized mapping network $\theta$ consistently outperforms the baseline. We believe this is due to the fact that adding a Gaussian distribution to an underlying latent distribution of StyleGAN is beneficial in smoothing out loss landscape, making it easier to perform OBI. Similar random projection can also be found in \cite{wu2019deep}.
\begin{table}
\centering
\begin{tabular}{@{}lccccc@{}}
\toprule
            & Number of Steps    & Full Model & Without CD and Buffer & Random $\theta$ & Baseline\\ \midrule
In-distribution & 20        & \textbf{3.082}      & 3.583      & 4.417      & 4.456                        \\
OOD             & 20        & \textbf{4.932}      & 5.127      & 5.135      & 5.292                        \\
In-distribution & 200       & \textbf{1.964}      & 3.034      & 2.723      & 2.569                        \\
OOD             & 200       & \textbf{3.498}      & 4.756      & 3.823      & 4.617                        \\
\bottomrule
\newline
\end{tabular}
\caption{\textbf{Ablation Study on Mapping Network.} Number of Steps indicates the number of optimization steps performed during inference. Evaluation metric consistent with \ref{fig:quantitative}.}
\label{tab:ablation}
\vspace{-2em}
\end{table}
\vspace{-1.2em}
\section{Conclusion}
This paper presents a method to accelerate optimization-based inference to invert a forward model. We propose an approach that learns a new space that is easier than the original input space to optimize with gradient descent at testing time. Our experiments and analysis on three different applications in computer  vision have shown that by learning this mapping function, optimization becomes more efficient and generalizes better to out-of-distribution data. Through quantitative and qualitative analysis, we found that such improvement in optimization performance comes from a smoother loss landscape. Since optimization-based inference has many advantages over encoder-based inference, we believe methods to accelerate them will have many impacts in a variety of applications.

\textbf{Acknowledgements:} This research is based on work partially supported by the NSF NRI Award \#1925157, NSF STC LEAP, the DARPA MCS program, and the DARPA CCU program. The views and conclusions contained herein are those of the authors and should not be interpreted as necessarily representing the official policies, either expressed or implied, of the sponsors.

\bibliography{reference}

\begin{thebibliography}{10}

\bibitem{abdal2019image2stylegan}
Rameen Abdal, Yipeng Qin, and Peter Wonka.
\newblock Image2stylegan: How to embed images into the stylegan latent space?
\newblock In {\em Proceedings of the IEEE/CVF International Conference on
  Computer Vision}, pages 4432--4441, 2019.

\bibitem{abdal2020image2stylegan}
Rameen Abdal, Yipeng Qin, and Peter Wonka.
\newblock Image2stylegan++: How to edit the embedded images?
\newblock In {\em Proceedings of the IEEE/CVF conference on computer vision and
  pattern recognition}, pages 8296--8305, 2020.

\bibitem{Abdal_2020_CVPR}
Rameen Abdal, Yipeng Qin, and Peter Wonka.
\newblock Image2stylegan++: How to edit the embedded images?
\newblock In {\em Proceedings of the IEEE/CVF Conference on Computer Vision and
  Pattern Recognition (CVPR)}, June 2020.

\bibitem{poseprior}
Ijaz Akhter and Michael Black.
\newblock Pose-conditioned joint angle limits for 3d human pose reconstruction.
\newblock 06 2015.

\bibitem{icnn}
Brandon Amos, Lei Xu, and J~Zico Kolter.
\newblock Input convex neural networks.
\newblock In {\em International Conference on Machine Learning}, pages
  146--155. PMLR, 2017.

\bibitem{bau2019seeing}
David Bau, Jun-Yan Zhu, Jonas Wulff, William Peebles, Hendrik Strobelt, Bolei
  Zhou, and Antonio Torralba.
\newblock Seeing what a gan cannot generate.
\newblock In {\em Proceedings of the IEEE/CVF International Conference on
  Computer Vision}, pages 4502--4511, 2019.

\bibitem{SGD}
L{\'e}on Bottou.
\newblock Large-scale machine learning with stochastic gradient descent.
\newblock In {\em Proceedings of COMPSTAT'2010}, pages 177--186. Springer,
  2010.

\bibitem{brakel2013training}
Phil{\'e}mon Brakel, Dirk Stroobandt, and Benjamin Schrauwen.
\newblock Training energy-based models for time-series imputation.
\newblock {\em The Journal of Machine Learning Research}, 14(1):2771--2797,
  2013.

\bibitem{CW}
Nicholas Carlini and David~A. Wagner.
\newblock Towards evaluating the robustness of neural networks.
\newblock In {\em 2017 {IEEE} Symposium on Security and Privacy}, pages 39--57,
  2017.

\bibitem{unlabeled}
Yair Carmon, Aditi Raghunathan, Ludwig Schmidt, John~C Duchi, and Percy~S
  Liang.
\newblock Unlabeled data improves adversarial robustness.
\newblock In H.~Wallach, H.~Larochelle, A.~Beygelzimer, F.~d\textquotesingle
  Alch\'{e}-Buc, E.~Fox, and R.~Garnett, editors, {\em Advances in Neural
  Information Processing Systems}, volume~32. Curran Associates, Inc., 2019.

\bibitem{chen2021full}
Boyuan Chen, Robert Kwiatkowski, Carl Vondrick, and Hod Lipson.
\newblock Full-body visual self-modeling of robot morphologies.
\newblock {\em arXiv preprint arXiv:2111.06389}, 2021.

\bibitem{cmumocap}
CMU.
\newblock Cmu mocap dataset.

\bibitem{cremer2019optimization}
Jochen~L Cremer, Ioannis Konstantelos, and Goran Strbac.
\newblock From optimization-based machine learning to interpretable security
  rules for operation.
\newblock {\em IEEE Transactions on Power Systems}, 34(5):3826--3836, 2019.

\bibitem{AA}
Francesco Croce and Matthias Hein.
\newblock Reliable evaluation of adversarial robustness with an ensemble of
  diverse parameter-free attacks.
\newblock In {\em ICML}, 2020.

\bibitem{Crowson2022VQGANCLIPOD}
Katherine Crowson, Stella~Rose Biderman, Daniel Kornis, Dashiell Stander, Eric
  Hallahan, Louis Castricato, and Edward Raff.
\newblock Vqgan-clip: Open domain image generation and editing with natural
  language guidance.
\newblock {\em ArXiv}, abs/2204.08583, 2022.

\bibitem{dinh2021hyperinverter}
Tan~M Dinh, Anh~Tuan Tran, Rang Nguyen, and Binh-Son Hua.
\newblock Hyperinverter: Improving stylegan inversion via hypernetwork.
\newblock {\em arXiv preprint arXiv:2112.00719}, 2021.

\bibitem{domke2012generic}
Justin Domke.
\newblock Generic methods for optimization-based modeling.
\newblock In {\em Artificial Intelligence and Statistics}, pages 318--326.
  PMLR, 2012.

\bibitem{goodfellow2014generative}
Ian Goodfellow, Jean Pouget-Abadie, Mehdi Mirza, Bing Xu, David Warde-Farley,
  Sherjil Ozair, Aaron Courville, and Yoshua Bengio.
\newblock Generative adversarial nets.
\newblock {\em Advances in neural information processing systems}, 27, 2014.

\bibitem{hassan2019resolving}
Mohamed Hassan, Vasileios Choutas, Dimitrios Tzionas, and Michael~J Black.
\newblock Resolving 3d human pose ambiguities with 3d scene constraints.
\newblock In {\em Proceedings of the IEEE/CVF international conference on
  computer vision}, pages 2282--2292, 2019.

\bibitem{hernandez2008multiview}
Carlos Hernandez, George Vogiatzis, and Roberto Cipolla.
\newblock Multiview photometric stereo.
\newblock {\em IEEE Transactions on Pattern Analysis and Machine Intelligence},
  30(3):548--554, 2008.

\bibitem{huh2020transforming}
Minyoung Huh, Richard Zhang, Jun-Yan Zhu, Sylvain Paris, and Aaron Hertzmann.
\newblock Transforming and projecting images into class-conditional generative
  networks.
\newblock In {\em European Conference on Computer Vision}, pages 17--34.
  Springer, 2020.

\bibitem{human36m}
Catalin Ionescu, Dragos Papava, Vlad Olaru, and Cristian Sminchisescu.
\newblock Human3.6m: Large scale datasets and predictive methods for 3d human
  sensing in natural environments.
\newblock {\em IEEE Transactions on Pattern Analysis and Machine Intelligence},
  36(7):1325--1339, 2014.

\bibitem{jahanian2019steerability}
Ali Jahanian, Lucy Chai, and Phillip Isola.
\newblock On the" steerability" of generative adversarial networks.
\newblock {\em arXiv preprint arXiv:1907.07171}, 2019.

\bibitem{jing2019neural}
Yongcheng Jing, Yezhou Yang, Zunlei Feng, Jingwen Ye, Yizhou Yu, and Mingli
  Song.
\newblock Neural style transfer: A review.
\newblock {\em IEEE transactions on visualization and computer graphics},
  26(11):3365--3385, 2019.

\bibitem{kang2021gan}
Kyoungkook Kang, Seongtae Kim, and Sunghyun Cho.
\newblock Gan inversion for out-of-range images with geometric transformations.
\newblock In {\em Proceedings of the IEEE/CVF International Conference on
  Computer Vision}, pages 13941--13949, 2021.

\bibitem{karras2017progressive}
Tero Karras, Timo Aila, Samuli Laine, and Jaakko Lehtinen.
\newblock Progressive growing of gans for improved quality, stability, and
  variation.
\newblock {\em arXiv preprint arXiv:1710.10196}, 2017.

\bibitem{karras2019style}
Tero Karras, Samuli Laine, and Timo Aila.
\newblock A style-based generator architecture for generative adversarial
  networks.
\newblock In {\em Proceedings of the IEEE/CVF conference on computer vision and
  pattern recognition}, pages 4401--4410, 2019.

\bibitem{karras2020analyzing}
Tero Karras, Samuli Laine, Miika Aittala, Janne Hellsten, Jaakko Lehtinen, and
  Timo Aila.
\newblock Analyzing and improving the image quality of stylegan.
\newblock In {\em Proceedings of the IEEE/CVF conference on computer vision and
  pattern recognition}, pages 8110--8119, 2020.

\bibitem{kingma2013auto}
Diederik~P Kingma and Max Welling.
\newblock Auto-encoding variational bayes.
\newblock {\em arXiv preprint arXiv:1312.6114}, 2013.

\bibitem{lee1993shape}
Kyoung~Mu Lee and C-CJ Kuo.
\newblock Shape from shading with a linear triangular element surface model.
\newblock {\em IEEE Transactions on Pattern Analysis and Machine Intelligence},
  15(8):815--822, 1993.

\bibitem{Lester2021ThePO}
Brian Lester, Rami Al-Rfou, and Noah Constant.
\newblock The power of scale for parameter-efficient prompt tuning.
\newblock {\em ArXiv}, abs/2104.08691, 2021.

\bibitem{shadows}
Ruoshi Liu, Sachit Menon, Chengzhi Mao, Dennis Park, Simon Stent, and Carl
  Vondrick.
\newblock Shadows shed light on 3d objects.
\newblock {\em arXiv}, 2022.

\bibitem{Liu2021FuseDreamTT}
Xingchao Liu, Chengyue Gong, Lemeng Wu, Shujian Zhang, Haoran Su, and Qiang
  Liu.
\newblock Fusedream: Training-free text-to-image generation with improved
  clip+gan space optimization.
\newblock {\em ArXiv}, abs/2112.01573, 2021.

\bibitem{mosh}
Matthew Loper, Naureen Mahmood, and Michael~J. Black.
\newblock Mosh: Motion and shape capture from sparse markers.
\newblock {\em ACM Trans. Graph.}, 33(6), nov 2014.

\bibitem{loper2015smpl}
Matthew Loper, Naureen Mahmood, Javier Romero, Gerard Pons-Moll, and Michael~J
  Black.
\newblock Smpl: A skinned multi-person linear model.
\newblock {\em ACM transactions on graphics (TOG)}, 34(6):1--16, 2015.

\bibitem{madry2017towards}
Aleksander Madry, Aleksandar Makelov, Ludwig Schmidt, Dimitris Tsipras, and
  Adrian Vladu.
\newblock Towards deep learning models resistant to adversarial attacks.
\newblock {\em arXiv preprint arXiv:1706.06083}, 2017.

\bibitem{mao2021adversarial}
Chengzhi Mao, Mia Chiquier, Hao Wang, Junfeng Yang, and Carl Vondrick.
\newblock Adversarial attacks are reversible with natural supervision.
\newblock In {\em Proceedings of the IEEE/CVF International Conference on
  Computer Vision}, pages 661--671, 2021.

\bibitem{TLA}
Chengzhi Mao, Ziyuan Zhong, Junfeng Yang, Carl Vondrick, and Baishakhi Ray.
\newblock Metric learning for adversarial robustness.
\newblock In {\em Advances in Neural Information Processing Systems},
  volume~32. Curran Associates, Inc., 2019.

\bibitem{menon2020pulse}
Sachit Menon, Alexandru Damian, Shijia Hu, Nikhil Ravi, and Cynthia Rudin.
\newblock Pulse: Self-supervised photo upsampling via latent space exploration
  of generative models.
\newblock In {\em Proceedings of the ieee/cvf conference on computer vision and
  pattern recognition}, pages 2437--2445, 2020.

\bibitem{mnih2013playing}
Volodymyr Mnih, Koray Kavukcuoglu, David Silver, Alex Graves, Ioannis
  Antonoglou, Daan Wierstra, and Martin Riedmiller.
\newblock Playing atari with deep reinforcement learning.
\newblock {\em arXiv preprint arXiv:1312.5602}, 2013.

\bibitem{pan2021exploiting}
Xingang Pan, Xiaohang Zhan, Bo~Dai, Dahua Lin, Chen~Change Loy, and Ping Luo.
\newblock Exploiting deep generative prior for versatile image restoration and
  manipulation.
\newblock {\em IEEE Transactions on Pattern Analysis and Machine Intelligence},
  2021.

\bibitem{pavlakos2019expressive}
Georgios Pavlakos, Vasileios Choutas, Nima Ghorbani, Timo Bolkart, Ahmed~AA
  Osman, Dimitrios Tzionas, and Michael~J Black.
\newblock Expressive body capture: 3d hands, face, and body from a single
  image.
\newblock In {\em Proceedings of the IEEE/CVF conference on computer vision and
  pattern recognition}, pages 10975--10985, 2019.

\bibitem{perarnau2016invertible}
Guim Perarnau, Joost Van De~Weijer, Bogdan Raducanu, and Jose~M {\'A}lvarez.
\newblock Invertible conditional gans for image editing.
\newblock {\em arXiv preprint arXiv:1611.06355}, 2016.

\bibitem{pidhorskyi2020adversarial}
Stanislav Pidhorskyi, Donald~A Adjeroh, and Gianfranco Doretto.
\newblock Adversarial latent autoencoders.
\newblock In {\em Proceedings of the IEEE/CVF Conference on Computer Vision and
  Pattern Recognition}, pages 14104--14113, 2020.

\bibitem{Radford2021LearningTV}
Alec Radford, Jong~Wook Kim, Chris Hallacy, Aditya Ramesh, Gabriel Goh,
  Sandhini Agarwal, Girish Sastry, Amanda Askell, Pamela Mishkin, Jack Clark,
  Gretchen Krueger, and Ilya Sutskever.
\newblock Learning transferable visual models from natural language
  supervision.
\newblock In {\em ICML}, 2021.

\bibitem{rice2020overfitting}
Leslie Rice, Eric Wong, and J.~Zico Kolter.
\newblock Overfitting in adversarially robust deep learning, 2020.

\bibitem{richardson2021encoding}
Elad Richardson, Yuval Alaluf, Or~Patashnik, Yotam Nitzan, Yaniv Azar, Stav
  Shapiro, and Daniel Cohen-Or.
\newblock Encoding in style: a stylegan encoder for image-to-image translation.
\newblock In {\em Proceedings of the IEEE/CVF Conference on Computer Vision and
  Pattern Recognition}, pages 2287--2296, 2021.

\bibitem{shen2020interpreting}
Yujun Shen, Jinjin Gu, Xiaoou Tang, and Bolei Zhou.
\newblock Interpreting the latent space of gans for semantic face editing.
\newblock In {\em Proceedings of the IEEE/CVF Conference on Computer Vision and
  Pattern Recognition}, pages 9243--9252, 2020.

\bibitem{stoyanov2011empirical}
Veselin Stoyanov, Alexander Ropson, and Jason Eisner.
\newblock Empirical risk minimization of graphical model parameters given
  approximate inference, decoding, and model structure.
\newblock In {\em Proceedings of the Fourteenth International Conference on
  Artificial Intelligence and Statistics}, pages 725--733. JMLR Workshop and
  Conference Proceedings, 2011.

\bibitem{intriguing}
Christian Szegedy, Wojciech Zaremba, Ilya Sutskever, Joan Bruna, Dumitru Erhan,
  Ian~J. Goodfellow, and Rob Fergus.
\newblock Intriguing properties of neural networks.
\newblock {\em arXiv:1312.6199}, 2013.

\bibitem{taheri2020grab}
Omid Taheri, Nima Ghorbani, Michael~J Black, and Dimitrios Tzionas.
\newblock Grab: A dataset of whole-body human grasping of objects.
\newblock In {\em European conference on computer vision}, pages 581--600.
  Springer, 2020.

\bibitem{tov2021designing}
Omer Tov, Yuval Alaluf, Yotam Nitzan, Or~Patashnik, and Daniel Cohen-Or.
\newblock Designing an encoder for stylegan image manipulation.
\newblock {\em ACM Transactions on Graphics (TOG)}, 40(4):1--14, 2021.

\bibitem{tripp2020sample}
Austin Tripp, Erik Daxberger, and Jos{\'e}~Miguel Hern{\'a}ndez-Lobato.
\newblock Sample-efficient optimization in the latent space of deep generative
  models via weighted retraining.
\newblock {\em Advances in Neural Information Processing Systems},
  33:11259--11272, 2020.

\bibitem{wang2019bidirectional}
Hao Wang, Chengzhi Mao, Hao He, Mingmin Zhao, Tommi~S Jaakkola, and Dina
  Katabi.
\newblock Bidirectional inference networks: A class of deep bayesian networks
  for health profiling.
\newblock In {\em Proceedings of the AAAI Conference on Artificial
  Intelligence}, volume~33, pages 766--773, 2019.

\bibitem{MART}
Yisen Wang, Difan Zou, Jinfeng Yi, James Bailey, Xingjun Ma, and Quanquan Gu.
\newblock Improving adversarial robustness requires revisiting misclassified
  examples.
\newblock In {\em ICLR}, 2020.

\bibitem{Wei2021ASB}
Tianyi Wei, Dongdong Chen, Wenbo Zhou, Jing Liao, Weiming Zhang, Lu~Yuan, Gang
  Hua, and Nenghai Yu.
\newblock A simple baseline for stylegan inversion.
\newblock {\em ArXiv}, abs/2104.07661, 2021.

\bibitem{welling2011bayesian}
Max Welling and Yee~W Teh.
\newblock Bayesian learning via stochastic gradient langevin dynamics.
\newblock In {\em Proceedings of the 28th international conference on machine
  learning (ICML-11)}, pages 681--688. Citeseer, 2011.

\bibitem{AWP}
Dongxian Wu, Shu-Tao Xia, and Yisen Wang.
\newblock Adversarial weight perturbation helps robust generalization.
\newblock In {\em NeurIPS}, 2020.

\bibitem{wu2019deep}
Yan Wu, Mihaela Rosca, and Timothy Lillicrap.
\newblock Deep compressed sensing.
\newblock In {\em International Conference on Machine Learning}, pages
  6850--6860. PMLR, 2019.

\bibitem{yeh2017semantic}
Raymond~A Yeh, Chen Chen, Teck Yian~Lim, Alexander~G Schwing, Mark
  Hasegawa-Johnson, and Minh~N Do.
\newblock Semantic image inpainting with deep generative models.
\newblock In {\em Proceedings of the IEEE conference on computer vision and
  pattern recognition}, pages 5485--5493, 2017.

\bibitem{yu2021plenoxels}
Alex Yu, Sara Fridovich-Keil, Matthew Tancik, Qinhong Chen, Benjamin Recht, and
  Angjoo Kanazawa.
\newblock Plenoxels: Radiance fields without neural networks.
\newblock {\em arXiv preprint arXiv:2112.05131}, 2021.

\bibitem{yu2015lsun}
Fisher Yu, Ari Seff, Yinda Zhang, Shuran Song, Thomas Funkhouser, and Jianxiong
  Xiao.
\newblock Lsun: Construction of a large-scale image dataset using deep learning
  with humans in the loop.
\newblock {\em arXiv preprint arXiv:1506.03365}, 2015.

\bibitem{zhang2021unsupervised}
Junzhe Zhang, Xinyi Chen, Zhongang Cai, Liang Pan, Haiyu Zhao, Shuai Yi,
  Chai~Kiat Yeo, Bo~Dai, and Chen~Change Loy.
\newblock Unsupervised 3d shape completion through gan inversion.
\newblock In {\em Proceedings of the IEEE/CVF Conference on Computer Vision and
  Pattern Recognition}, pages 1768--1777, 2021.

\bibitem{zhang2018unreasonable}
Richard Zhang, Phillip Isola, Alexei~A Efros, Eli Shechtman, and Oliver Wang.
\newblock The unreasonable effectiveness of deep features as a perceptual
  metric.
\newblock In {\em Proceedings of the IEEE conference on computer vision and
  pattern recognition}, pages 586--595, 2018.

\bibitem{zhu2020domain}
Jiapeng Zhu, Yujun Shen, Deli Zhao, and Bolei Zhou.
\newblock In-domain gan inversion for real image editing.
\newblock In {\em European conference on computer vision}, pages 592--608.
  Springer, 2020.

\bibitem{zhu2016generative}
Jun-Yan Zhu, Philipp Kr{\"a}henb{\"u}hl, Eli Shechtman, and Alexei~A Efros.
\newblock Generative visual manipulation on the natural image manifold.
\newblock In {\em European conference on computer vision}, pages 597--613.
  Springer, 2016.

\end{thebibliography}

\appendix

\newpage

\section{Algorithm}
Accompanying Section \ref{training}, we provide a detailed coordinate ascent training algorithm with an experience replay buffer:

\begin{algorithm}
\caption{Learning Mapping Network $\theta$}
\label{algorithm: defense}
\begin{algorithmic}[1]
\STATE {\bfseries Input:} Ground truth $y$, step size $\lambda_z$ and $\lambda_{\theta}$, number of buffers $B$, number of data samples in a buffer $N$, number of optimization steps for each sample $T$, loss function $L$, and forward model $F$.
\STATE {\bfseries Output:} Learned mapping network $\theta$
\STATE{Randomly initialize a mapping network $\theta$}
\FOR{$b=1,...,B$}
\STATE{Initialize Experience Replay Buffer $[\{\z_{t, i}\}_{t = 1} ^ {T}]_{i=1}^{N}$}
\FOR{$i=1,...,N$}
\STATE{$\z_0 \leftarrow \0$}
\FOR{$t=1,...,T$}
\STATE{$l \leftarrow L(F(\theta(\z)))$}
\STATE{$\z_{t} \leftarrow \z_{t-1} + \lambda_z \frac{\partial l}{\partial \z_{t-1}}$}
\STATE{$\z_{t, i} \leftarrow z_{t}$}
\ENDFOR
\ENDFOR
\FOR{$j=1,...,T \cdot N$}
\STATE{Randomly sample $\z$ from previously collected buffer}
\STATE{$l \leftarrow L(F(\theta(\z)))$}
\STATE{$\theta_j \leftarrow \theta_{j-1} + \lambda_\theta \frac{\partial L}{\partial \theta_{j-1}}$}
\ENDFOR

\ENDFOR
\STATE{Return $\theta$}
\end{algorithmic}
\label{alg:method}
\end{algorithm}

\section{Implementation Details}
We will released all code, models, and data. Here we describe our implementation details for the above algorithm.
\subsection{GAN Inversion}
Mapping network $\theta$ is implemented with a 3-layer MLP. The input dimension (dimension of $Z$ space) is the same as the output dimension (dimension of $X$ space). For each intermediate output, we apply a Leaky ReLu function with a negative slope of 0.2 as an activation function. Hidden dimension of the MLP is 1024. For optimizing $z$ (collecting optimization trajectories), we use an Adam optimizer with a learning rate of 0.1. For training the mapping network $\theta$, we use an AdamW optimizer with a weight decay of 0.1 with a learning rate of 0.0001. We used the following parameter set, $T = 20$, $N = 256$, $B = 500$.  For baseline, we use the implementation of \cite{abdal2019image2stylegan} provided \href{https://github.com/zaidbhat1234/Image2StyleGAN}{here}. The pretrained weights of StyleGAN converted to PyTorch are also provided in the same link.

\subsection{3D Human Pose Reconstruction}
Mapping network $\theta$ is implemented with a 3-layer MLP. The input dimension is 128 (dimension of $Z$) and the output dimension is 32 (dimension of $X$). For each intermediate output, we apply a Leaky ReLu function with a negative slope of 0.2 as an activation function. Hidden dimension of the MLP is 512. For optimizing $z$ (collecting optimization trajectories), we use an Adam optimizer with a learning rate of 0.1. For training the mapping network $\theta$, we use an AdamW optimizer with a weight decay of 0.1 and a learning rate of 0.005. We use the following parameter set, $T = 200$, $N = 40960$, $B = 500$. For a fair comparison with the baseline, we tried a range of learning rate for $\x \in X$ $\{0.5, 0.1, 0.05, 0.01, 0.001\}$ and select the best performing configuration for comparison.

\subsection{Defending Adversarial Attacks}
Mapping network $\theta$ is implemented with a 3-layer MLP. The input dimension is 3072 (dimension of $Z$) and the output dimension is 3072 (dimension of $X$). For each intermediate output, we apply a Leaky ReLu function with a negative slope of 0.2 as an activation function. Hidden dimension of the MLP is 3072. For optimizing $z$ (collecting optimization trajectories), we use an Adam optimizer with learning rate $0.2/255$. For training the mapping network $\theta$, we use an AdamW optimizer with learning rate of 0.0001 and a weight decay of 0.1.
 We use the following parameter set, $T = 5$, $N = 5120$, $B = 70$.
For the regularization term $\lambda$ that constrains the amplitude of the additive defense vector, we use $\lambda=1$. We use random start instead of zero start for initializing the attack reversal vector.

%  $\lambda_z = 0.2/255$, $\lambda_\theta = 1e-4$,

\section{Limitations}

Optimization-based inference has intrinsic advantages to robustness, accuracy, and flexibility, which comes at the cost of additional computation time during inference. Encoder-based methods will usually perform faster because they only require a single forward pass of a neural network, while our approach requires several computational passes in both the forward and backward (gradient) direction. We believe that for many applications this trade-off will be desirable, especially in cases where accuracy is more important than speed.  Our approach aims to minimize this additional computational overhead brought by optimization-based inference, and our experiments on multiple datasets show the significant computational savings compared to other optimization-based inference methods. 

Unlike many other optimization-based inference algorithms, our approach also requires a training step in order to fit a suitable landscape, which requires both training time and training data. However, we believe this overhead is insignificant for most applications and we have designed our neural networks to be efficient. For example, $\theta$ is relatively lightweight, making its training time fairly marginal compared to the training of the forward model $F$. In all our experiments, we found that the training time of $\theta$ is orders of magnitudes faster than the training time for $F$.

\section{Societal Impact}
Optimization-based inference has a wide variety of applications broadly in computer vision, robotics, natural language processing, assistive technology, security, and healthcare. Since our proposed method provides significant acceleration to these inference techniques, we expect our work to find positive impact in these applications, where speed, accuracy, and robustness are often critical.

Our algorithm is compatible with many different types of forward models -- as long as a gradient can be calculated -- including neural networks. However, learned forward models are known to acquire and contain biases from the original dataset. Our approach will also inherit the same biases. While our approach to inference offers some advantages to out-of-distribution data, it does not mitigate nor correct biases. The datasets in our experiments are likely not representative of the population, and consequently also contain biases. We acknowledge these limitations, and applications of this method should be mindful of these limitations, especially in potentially sensitive scenarios. As the community continues to address biases in models and datasets, our method can be applied to these models in the future too.

\end{document}